\documentclass[conference]{IEEEtran}
\IEEEoverridecommandlockouts

\usepackage{cite}
\usepackage{amsmath,amssymb,amsfonts}
\usepackage{alphabeta}
\usepackage{algorithmic}
\usepackage{graphicx}
\usepackage{textcomp}
\usepackage{xcolor}
\def\BibTeX{{\rm B\kern-.05em{\sc i\kern-.025em b}\kern-.08em
    T\kern-.1667em\lower.7ex\hbox{E}\kern-.125emX}}
\begin{document}

\title{A Dynamic Framework for Grid Adaptation in Kolmogorov--Arnold Networks}

\author{
	\IEEEauthorblockN{Spyros Rigas\IEEEauthorrefmark{1}, Thanasis Papaioannou\IEEEauthorrefmark{1}, Panagiotis Trakadas\IEEEauthorrefmark{2}, and Georgios Alexandridis\IEEEauthorrefmark{1}}
	\IEEEauthorblockA{\IEEEauthorrefmark{1}Department of Digital Industry Technologies, National and Kapodistrian University of Athens, Greece}
	\IEEEauthorblockA{\IEEEauthorrefmark{2}Department of Port Management and Shipping, National and Kapodistrian University of Athens, Greece}
	\IEEEauthorblockA{Email: \{spyrigas, atpapaioannou, ptrakadas, gealexandri\}@uoa.gr}
}


\maketitle

\begin{abstract}
Kolmogorov--Arnold Networks (KANs) have recently demonstrated promising potential in scientific machine learning, partly due to their capacity for grid adaptation during training. However, existing adaptation strategies rely solely on input data density, failing to account for the geometric complexity of the target function or metrics calculated during network training. In this work, we propose a generalized framework that treats knot allocation as a density estimation task governed by Importance Density Functions (IDFs), allowing training dynamics to determine grid resolution. We introduce a curvature-based adaptation strategy and evaluate it across synthetic function fitting, regression on a subset of the Feynman dataset and different instances of the Helmholtz PDE, demonstrating that it significantly outperforms the standard input-based baseline. Specifically, our method yields average relative error reductions of 25.3\% on synthetic functions, 9.4\% on the Feynman dataset, and 23.3\% on the PDE benchmark. Statistical significance is confirmed via Wilcoxon signed-rank tests, establishing curvature-based adaptation as a robust and computationally efficient alternative for KAN training.
\end{abstract}


\section{Introduction} \label{sec1}

Kolmogorov--Arnold Networks (KANs) \cite{KAN1} have recently emerged as a promising alternative to Multilayer Perceptrons (MLPs), mitigating several of the shortcomings of MLP-based architectures, such as spectral bias \cite{KAN3}, as well as demonstrating improved expressivity and interpretability \cite{GenReview}. These, along with other advantageous properties, have allowed architectures that use them as their primary backbone to find several applications in engineering and industrial domains \cite{Robots, CKAN, bearings, IntrusionDetection, Energy}. However, their most profound impact has arguably been observed in the domain of scientific machine learning \cite{KAN2}, with numerous applications and even state-of-the-art results in fields like Physics-Informed Machine Learning (PIML) \cite{FAIR, rigasAdaptive, KINN, rgakans, VlasovPoisson} and operator learning \cite{DeepOKAN, KANO}, among others \cite{KANODEs, PointNet, EqDiscovery, Earth}.

Following the introduction of the original spline-based architecture, referred to as ``vanilla'' KANs, various studies proposed replacing the B-spline basis with alternative, more computationally efficient functions, such as Chebyshev polynomials \cite{FAIR} or ReLU-based \cite{ReLUKAN} basis functions. While such approaches have seen success in specific problem settings \cite{GenReview, PINNReview}, the reliance on grid-independent basis functions precludes the possibility of fine-grained training via grid extension \cite{KAN1, rigasAdaptive}, i.e., the capacity to start training with a coarse grid and progressively expand it to introduce more grid points (knots). Moreover, even when the grid resolution is fixed, the ability to adaptively reallocate knots during training -- a process known as grid adaptation -- has been shown to improve accuracy compared to using static grids \cite{rigasAdaptive, adaptKAN}. To this end, in this paper we focus on ``fully adaptive'' basis functions (adopting the terminology introduced in \cite{rigasAdaptive}), specifically the standard B-splines.

Despite the benefits of full grid adaptivity, the strategies currently employed to determine knot allocation during grid adaptation remain largely restricted to input-based heuristics. In the seminal implementation \cite{KAN1}, the authors introduced an adaptive strategy where knots are placed according to the density of each layer's input points. Recently, \cite{adaptKAN} proposed the AdaptKAN framework to automate this process via moving histograms of the data; nonetheless, their solution also remains dependent on the input data distribution. In this work, we propose a broader framework for knot allocation that reduces to the aforementioned input-based strategies in special cases, but is also capable of utilizing training dynamics. This is achieved by defining Importance Density Functions (IDFs), which utilize metrics calculated during training on either a global or a per-layer basis. Although a diverse set of IDFs can be defined, in this work, we utilize the concept of curvature to guide knot allocation, drawing inspiration from classical B-spline approximation theory \cite{deboor}.

In particular, the main contributions of our work are the following:

\begin{itemize}
	\item We introduce a generalized framework for dynamic grid adaptation in KANs, demonstrating that the input-based method introduced in \cite{KAN1} can be obtained as the limiting case of a uniform IDF.
	
	\item We propose a curvature-based adaptation strategy to guide knot allocation during training.
	
	\item We evaluate our framework on function regression tasks using both a synthetic dataset of custom-curated functions and a subset of the Feynman dataset, as well as four instances of a forward PDE problem.
\end{itemize}

The remainder of this paper is organized as follows. In Section \ref{sec2}, we present the vanilla KAN layer, which is the foundational building block of the architectures used in our work. Section \ref{sec3} introduces our proposed framework, defining the concept of IDFs and deriving the input-based strategy of \cite{KAN1} as a special case. We also detail the specific curvature-based IDF utilized in this study. Section \ref{sec4} presents the experimental setup, followed by results across three distinct benchmarks: a suite of synthetic functions, a subset of the Feynman dataset \cite{Feynman} and different instances of a forward PDE problem. Finally, Section \ref{sec5} summarizes our key findings, discusses current limitations and outlines directions for future research. We note that, to support reproducibility, all source code and data utilized in this work are publicly available via a GitHub repository (see Reproducibility Statement).

\section{Kolmogorov--Arnold Networks} \label{sec2}

A (vanilla) KAN is a collection of layers where each layer's output, $\mathbf{y} \in \mathbb{R}^{n_\text{out}}$, is related to its input, $\mathbf{x} \in \mathbb{R}^{n_\text{in}}$, via:

\begin{equation}
	y_j = \sum_{i=1}^{n_\text{in}} 
	\left( r_{ji} \, R\left(x_i\right) + c_{ji} \sum_{m=1}^{G+k} b_{jim} \, B_m \left(x_i\right) \right), 
	\label{eq1}
\end{equation}

\noindent with $j \in \left\{ 1, \dots, n_\text{out}\right\}$. In this expression, 

\begin{equation}
	R\left(x\right) = \frac{x}{1 + \exp\left(-x\right)}
	\label{eq2}
\end{equation}

\noindent is the Sigmoid Linear Unit (SiLU) function, and $B_m\left( \cdot \right)$ are B-spline basis functions of order $k$. The construction of these basis functions relies on a knot vector $\mathbf{t}$, which is generated in two distinct steps. First, the input domain is partitioned into $G$ intervals; the distribution of these intervals determines the local resolution of the spline. Second, this primary partition is augmented by extending the vector at its endpoints to ensure the basis functions provide appropriate coverage at the domain boundaries \cite{KAN1}. 

Importantly, the framework proposed in this work focuses solely on the determination of the primary partition, as the subsequent augmentation is a straightforward, deterministic procedure. As indicated in Eq. (\ref{eq1}), for a grid of $G$ intervals and spline order $k$, the resulting basis contains $G+k$ trainable coefficients. Consequently, the representational power of the network is governed not only by the network depth and width, but also by the specific allocation of the knots in $\mathbf{t}$, which motivates our study. For the purposes of this work, the trainable scaling weights are initialized as $c_{ji} = 1$, following \cite{KAN1}, while the trainable residual and basis weights ($r_{ji}$ and $b_{jim}$) are initialized following the Glorot-inspired strategy introduced in \cite{initICLR}.

\section{Proposed Framework} \label{sec3}

We approach the problem of knot allocation as a density estimation task. Our objective is to determine a knot vector $\mathbf{t}$ for a given KAN layer such that the density of the grid points mirrors an underlying Importance Density Function (IDF), which characterizes the ``significance'' of specific regions within the layer's input domain. While the standard KAN implementation inherently assumes that importance is equivalent to the input data probability density function, we generalize this by allowing importance to be derived from the training dynamics of the network itself.

\subsection{Generalized Knot Allocation} \label{sec3.1}

Let $\{\mathbf{x}^{(s)}\}_{s=1}^{N_b}$ denote a batch of input coordinates incident to a specific layer, where $\mathbf{x}^{(s)} \in \mathbb{R}^{n_{\text{in}}}$. For a specific feature dimension $d \in \{1, \dots, n_{\text{in}}\}$, we seek to distribute knots based on a set of scalar weights $\{w^{(s)}\}_{s=1}^{N_b}$, where $w^{(s)}$ assigns a local importance score to the coordinate $x_d^{(s)}$.

We define the discrete empirical IDF as a probability mass function over the input samples:

\begin{equation}
	P\left(x_d^{(s)}\right) = \frac{w^{(s)}}{\sum_{j=1}^{N_b} w^{(j)}}.
	\label{eq3}
\end{equation}

\noindent To determine the knot positions, we construct the weighted empirical cumulative distribution function, denoted as $\hat{F}_d(z)$. This is a step function bounded in $[0, 1]$:

\begin{equation}
	\hat{F}_d(z) = \sum_{s=1}^{N_b} P\left(x_d^{(s)}\right) \cdot \mathbb{I}\left(x_d^{(s)} \le z\right),
	\label{eq4}
\end{equation}

\noindent where $\mathbb{I}(\cdot)$ is the indicator function. The knot vector $\mathbf{t}$ is obtained by inverting this distribution via the generalized inverse distribution function (quantile function):

\begin{equation}
	t_m = \hat{F}_d^{-1}(q_m) = \inf \{ z \in \mathbb{R} : \hat{F}_d(z) \ge q_m \},
	\label{eq5}
\end{equation}

\noindent where $q_m = m / (G+1)$ represents the target quantile for the $m$-th knot. In practice, this inversion is performed by sorting the samples along dimension $d$ such that $x_d^{(1)} \le \dots \le x_d^{(N_b)}$, computing the cumulative sum of their sorted weights, and selecting the coordinate $x_d^{(p)}$ where the cumulative mass first exceeds the threshold $q_m$.

\subsection{Input-Based Adaptation as a Special Case} \label{sec3.2}

The standard grid adaptation strategy employed in the original KAN implementation \cite{KAN1} allocates knots solely based on the input data distribution. The underlying rationale is that regions with high data density require finer resolution to minimize approximation error.

Within our proposed framework, this strategy is retrieved as a special case by selecting a uniform IDF, where the importance weights are constant for all samples:

\begin{equation}
	w^{(s)}_{\text{uniform}} = 1, \quad \forall s \in \{1, \dots, N_b\}.
	\label{eq6}
\end{equation}

\noindent Under this condition, the resulting grid has uniform mass with respect to the sample count, thus concentrating knots in regions of high input density. While effective for ensuring that basis functions are active where data exists, this approach ignores the complexity of the function being approximated, potentially allocating excessive resolution to linear regions simply because they are oversampled.

\subsection{Curvature-Based Adaptation} \label{sec3.3}

Classical spline approximation theory links optimal knot placement to the derivatives of the target function. Specifically, de Boor \cite{deboor} demonstrates that the asymptotic knot density should scale with $|f^{(k+1)}(x)|^{1/(k+1)}$. While for the cubic B-splines utilized in this work ($k=3$) this theoretically points to the fourth derivative, we adopt the curvature as our guiding metric. This choice is motivated by the fact that by allocating knots proportional to curvature, we ensure that the grid densifies in regions exhibiting significant geometric features, while rarefying in linear regions. This serves as a basic implementation of our framework, noting that it is straightforward to define a corresponding IDF for higher-order derivatives.

We estimate the local curvature at a sample point via the diagonal of the Hessian of the layer's response. Specifically, let $\Phi(\mathbf{x}): \mathbb{R}^{n_{\text{in}}} \to \mathbb{R}^{n_{\text{out}}}$ denote the mapping learned by the layer. When adapting the grid for the $d$-th input coordinate, we define the importance weight $w^{(s)}_{\text{curv}}$ as the aggregated second-order partial derivative along that axis:

\begin{equation}
	w^{(s)}_{\text{curv}} = \sum_{j=1}^{n_{\text{out}}} \left| \frac{\partial^2 \Phi_j}{\partial x_d^2} \left(\mathbf{x}^{(s)}\right) \right| + \epsilon,
	\label{eq7}
\end{equation}

\noindent where $\epsilon$ is a small smoothing constant to ensure strict positivity. In practice, Eq. \eqref{eq7} can be computed exactly in frameworks with automatic differentiation such as \texttt{JAX} \cite{JAX}, otherwise it can be approximated via central finite differences.

\subsection{An Illustrative Example} \label{sec3.4}

To demonstrate the difference between input-based and curvature-based adaptation, we consider a 1D regression task involving a highly localized feature: $f(x) = \exp\left( - 200 x^2 \right)$. The training dataset consists of $N=1000$ samples drawn from a uniform distribution over $[-1, 1]$. We train a KAN with architecture $[1, 10, 1]$ for 1000 training iterations using a coarse initial grid of $G=3$ intervals. At the final epoch, we perform grid extension, increasing the grid resolution to $G=10$ while simultaneously adapting the knot positions using the two strategies.

\begin{figure}[b!]
	\centerline{\includegraphics[width=\linewidth]{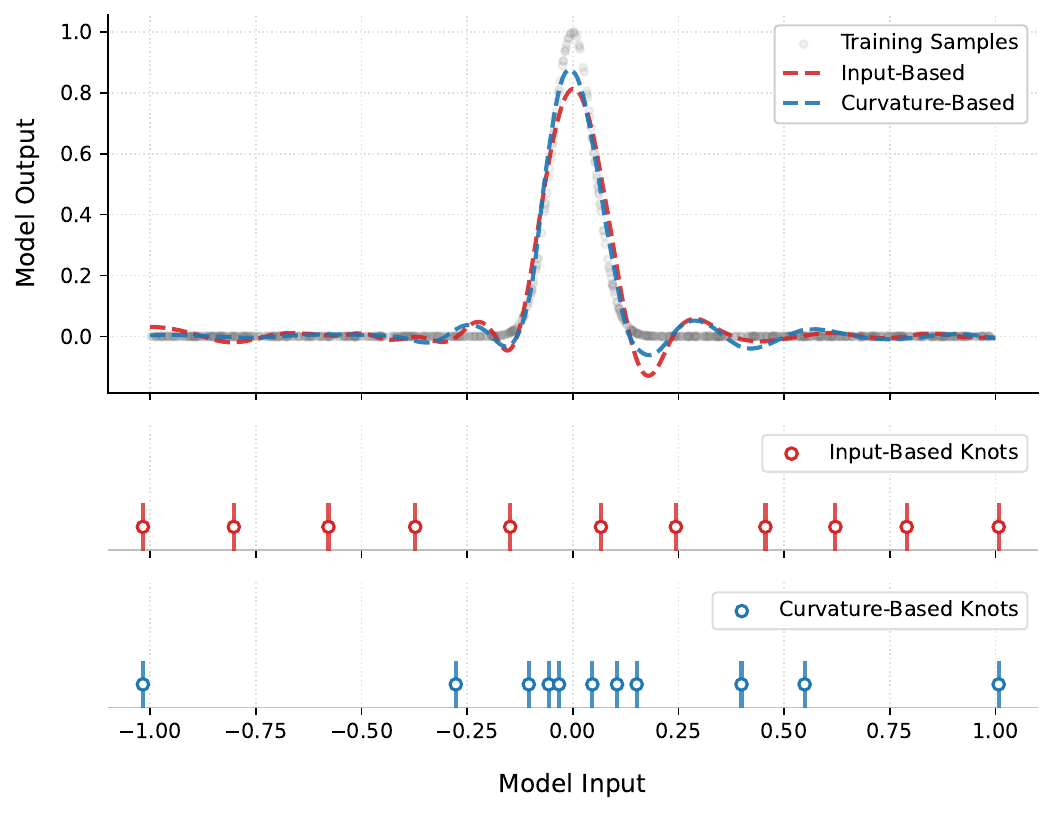}}
	\caption{Visual comparison of grid adaptation strategies on a 1D sharp Gaussian regression task. (Top) The model predictions from the input-based (dashed red) and curvature-based (dashed blue) KAN models. The training data, sampled from a uniform distribution, is shown in light gray. (Middle) The resulting knot allocation for the input-based strategy. (Bottom) The knot allocation for the proposed curvature-based strategy. }
	\label{fig:comparison}
\end{figure}

Figure \ref{fig:comparison} visualizes the outcome of this experiment. The top row displays the function approximation at the moment of grid update. The middle and bottom rows depict the knot allocation for the input-based and curvature-based strategies, respectively. As expected, the input-based strategy mirrors the uniform data distribution, resulting in an almost equidistant grid that wastes capacity on the linear ``tails'' of the Gaussian where $f(x) \approx 0$. Conversely, the curvature-based strategy identifies the region of high second-derivative magnitude. It condenses the knots near the peak ($x \approx 0$) and the inflection points, providing dense support exactly where the function requires high resolution, while rarefying the grid in the flat regions. We emphasize that this experiment serves primarily as a visual demonstration of the studied strategies; a quantitative evaluation follows in Section \ref{sec4}.

\section{Experimental Results} \label{sec4}

In this section, we present experiments across three distinct benchmarks: a suite of custom synthetic functions, a subset of the dimensionless Feynman dataset and four instances of the Helmholtz equation within the PIML setting \cite{PIML}. Across all benchmarks, we compare two primary configurations: (i) a baseline model using the standard input-based adaptation strategy (equivalent to uniform IDF), and (ii) a model employing the proposed curvature-based adaptation. To isolate the impact of knot allocation, all other hyperparameters remain identical between the two configurations, utilizing the Adam optimizer to minimize the Mean Squared Error (MSE) loss. Training is initiated with a coarse grid of $G=3$ intervals. We employ a progressive grid extension schedule, increasing the resolution to $G=6$, $G=9$, and finally $G=12$ at predefined intervals throughout the training process. At each extension step, the knot positions are reallocated according to the respective adaptation strategy.

To ensure the statistical robustness of our findings, we conduct 3 independent runs for each experiment using different random seeds. Performance is evaluated using the median relative $L^2$ error across the 3 runs with respect to the reference solution defined on a dense grid. Furthermore, to statistically validate our comparisons, we perform a one-tailed Wilcoxon signed-rank test \cite{wilcox}. The null hypothesis posits that the curvature-based strategy does not yield a statistically significant reduction in error compared to the input-based baseline. All experiments are conducted using the \texttt{jaxKAN} framework \cite{jaxkan} on a single NVIDIA GeForce RTX 4090 GPU. The source code and datasets used for this study will become available online upon acceptance of the manuscript.

\subsection{Benchmark I: Synthetic Functions} \label{sec4.1}

We first define a custom curated benchmark consisting of 10 synthetic functions (from 1 to 6 dimensions), selected to exhibit challenging characteristics such as highly localized features, sharp discontinuities and varying dimensionalities. This suite is designed to explicitly test whether the proposed curvature-based IDF can effectively allocate resolution to ``hard'' regions of the domain better than the input density.

For each function, we sample $N = 4000$ training points and train a KAN model with a hidden dimension of 10 for 2000 iterations using a constant learning rate of $\eta = 10^{-2}$. The progressive grid updates are performed at iterations 500, 1000 and 1500. Table \ref{tab1} presents the final relative $L^2$ error for both strategies. We also report the relative percent improvement provided by the curvature method, where positive values denote a reduction in error and negative values indicate deterioration.

\begin{table}[t!]
	\caption{Comparison of Relative $L^2$ Error on Synthetic Functions (Median $\pm$ Std over 3 seeds)}
	\begin{center}
		\renewcommand{\arraystretch}{1.2}
		\begin{tabular}{|c|c|c|c|}
			\hline
			\textbf{Function} & \textbf{Input-Based} & \textbf{Curvature-Based} & \textbf{Improv.} \\
			\hline
			$f_1$          & $(3.4 \pm 0.4) \cdot 10^{-2}$ & $\mathbf{(2.9 \pm 0.4) \cdot 10^{-2}}$ & +14.61\% \\ 
			$f_2$              & $(2.4 \pm 0.7) \cdot 10^{-3}$ & $\mathbf{(1.0 \pm 1.0) \cdot 10^{-3}}$ & +57.93\% \\ 
			$f_3$          & $(1.3 \pm 0.3) \cdot 10^{-2}$ & $\mathbf{(1.0 \pm 0.1) \cdot 10^{-2}}$ & +25.69\% \\ 
			$f_4$              & $(2.9 \pm 2.4) \cdot 10^{-3}$ & $\mathbf{(1.9 \pm 1.4) \cdot 10^{-3}}$ & +35.51\% \\ 
			$f_5$              & $(1.1 \pm 0.5) \cdot 10^{-2}$ & $\mathbf{(6.7 \pm 1.3) \cdot 10^{-3}}$ & +36.70\% \\ 
			$f_6$              & $(2.4 \pm 3.3) \cdot 10^{-3}$ & $\mathbf{(2.1 \pm 1.3) \cdot 10^{-3}}$ & +11.49\% \\ 
			$f_7$          & $(8.6 \pm 8.6) \cdot 10^{-4}$ & $\mathbf{(6.5 \pm 2.7) \cdot 10^{-4}}$ & +25.09\% \\ 
			$f_8$              & $(6.7 \pm 23.5) \cdot 10^{-3}$ & $\mathbf{(5.1 \pm 1.8) \cdot 10^{-3}}$ & +24.71\% \\ 
			$f_9$ & $(9.8 \pm 3.9) \cdot 10^{-3}$ & $\mathbf{(7.1 \pm 3.4) \cdot 10^{-3}}$ & +27.80\% \\ 
			$f_{10}$    & $\mathbf{(8.1 \pm 5.6) \cdot 10^{-2}}$ & $(8.6 \pm 3.7) \cdot 10^{-2}$ & -6.42\% \\ 
			\hline
		\end{tabular}
		\label{tab1}
	\end{center}
\end{table}

The results indicate that the curvature-based strategy outperforms the input-based baseline in 9 out of 10 cases, yielding an average relative improvement of 25.31\% and reaching a maximum error reduction of over 50\% for function $f_2$. Notably, the curvature-based results generally exhibit lower standard deviations compared to the baseline, suggesting that the proposed method offers not only higher accuracy but also improved consistency across independent runs. The one-tailed Wilcoxon signed-rank test yields a $p$-value of 0.042; since $p < 0.05$, we reject the null hypothesis, concluding that the curvature-based method provides a statistically significant reduction in approximation error. Regarding computational cost, the average wall-clock time per run is 15.47s for the input-based method compared to 17.13s for the curvature-based approach. This approximate 10\% increase is to be expected, since the IDF in the curvature-based case is more complex than the uniform IDF of the input-based one (compare Eq. \eqref{eq7} to Eq. \eqref{eq6}), and can be considered a negligible trade-off given the substantial gains in predictive performance.

\subsection{Benchmark II: Feynman Dataset} \label{sec4.2}

While the previous benchmark was specifically curated to highlight the advantages of curvature-based adaptation on functions with a certain geometric complexity, a robust knot allocation strategy should, at minimum, match the performance of the baseline on generic regression tasks. To evaluate this generalizability, we turn to a subset of the dimensionless Feynman dataset \cite{Feynman}, consisting of 15 physics equations which have been utilized in prior KAN literature \cite{KAN1, initICLR}.

For this benchmark, we employ a KAN architecture with a hidden dimension of 10. Training is conducted for 2000 iterations using a learning rate of $\eta = 10^{-3}$. Consistent with the previous experiment, grid extension and adaptation steps are performed at iterations 500, 1000 and 1500. The comparative results in terms of final relative $L^2$ error are summarized in Table \ref{tab2}.

\begin{table}[t!]
	\caption{Comparison of Relative $L^2$ Error on Feynman Dataset \protect\\(Median $\pm$ Std over 3 seeds)}
	\begin{center}
		\renewcommand{\arraystretch}{1.2}
		\begin{tabular}{|c|c|c|c|}
			\hline
			\textbf{Index} & \textbf{Input-Based} & \textbf{Curvature-Based} & \textbf{Improv.} \\
			\hline
			I.6.2  & $(6.0 \pm 1.3) \cdot 10^{-1}$ & $\mathbf{(5.5 \pm 1.6) \cdot 10^{-1}}$ & +8.48\% \\ 
			I.12.11  & $(3.4 \pm 1.0) \cdot 10^{-4}$ & $\mathbf{(3.2 \pm 2.7) \cdot 10^{-4}}$ & +7.74\% \\ 
			I.13.12  & $(9.4 \pm 0.5) \cdot 10^{-1}$ & $\mathbf{(7.5 \pm 1.4) \cdot 10^{-1}}$ & +20.68\% \\ 
			I.16.6  & $(2.2 \pm 0.1) \cdot 10^{-2}$ & $\mathbf{(2.2 \pm 0.0) \cdot 10^{-2}}$ & +0.81\% \\ 
			I.18.4  & $\mathbf{(1.0 \pm 0.0) \cdot 10^{0}}$ & $\mathbf{(1.0 \pm 0.0) \cdot 10^{0}}$ & 0.00\% \\ 
			I.27.6  & $\mathbf{(1.0 \pm 0.0) \cdot 10^{0}}$ & $\mathbf{(1.0 \pm 0.0) \cdot 10^{0}}$ & 0.00\% \\ 
			I.29.16  & $(5.6 \pm 1.5) \cdot 10^{-3}$ & $\mathbf{(5.4 \pm 0.8) \cdot 10^{-3}}$ & +4.08\% \\ 
			I.30.3  & $(5.5 \pm 1.3) \cdot 10^{-4}$ & $\mathbf{(5.2 \pm 1.7) \cdot 10^{-4}}$ & +5.18\% \\ 
			I.40.1  & $(9.0 \pm 1.7) \cdot 10^{-4}$ & $\mathbf{(7.4 \pm 3.0) \cdot 10^{-4}}$ & +18.40\% \\ 
			II.2.42 & $(5.1 \pm 2.9) \cdot 10^{-4}$ & $\mathbf{(4.4 \pm 1.5) \cdot 10^{-4}}$ & +14.36\% \\ 
			II.6.15a & $(1.8 \pm 0.7) \cdot 10^{-2}$ & $\mathbf{(1.7 \pm 0.7) \cdot 10^{-2}}$ & +8.94\% \\ 
			II.11.7 & $(2.2 \pm 0.3) \cdot 10^{-3}$ & $\mathbf{(1.9 \pm 0.4) \cdot 10^{-3}}$ & +10.66\% \\ 
			II.35.18 & $(4.5 \pm 2.1) \cdot 10^{-4}$ & $\mathbf{(3.7 \pm 6.7) \cdot 10^{-4}}$ & +17.51\% \\ 
			II.36.38 & $(2.1 \pm 0.5) \cdot 10^{-3}$ & $\mathbf{(1.8 \pm 0.6) \cdot 10^{-3}}$ & +14.81\% \\ 
			III.17.37 & $(2.3 \pm 0.3) \cdot 10^{-3}$ & $\mathbf{(2.1 \pm 0.3) \cdot 10^{-3}}$ & +9.89\% \\ 
			\hline
		\end{tabular}
		\label{tab2}
	\end{center}
\end{table}

The results indicate that the proposed curvature-based strategy consistently outperforms the input-based baseline, achieving a lower median error in 13 out of 15 test cases. The two exceptions (I.18.4 and I.27.6) represent cases where both strategies fail to converge to a low-error solution, resulting in negligible differences in performance. While the improvement is less pronounced than in the synthetic benchmark, yielding an average reduction of 9.44\% and a maximum of approximately 20\% for function  I.13.12, this is expected, as physical laws generally yield smoother, well-behaved functions compared to the synthetic examples considered in Benchmark I.

Nonetheless, the consistency of these gains results in strong statistical significance. The one-tailed Wilcoxon signed-rank test yields a $p$-value of $1.5 \cdot 10^{-4}$ ($p \ll 0.05$), leading to a decisive rejection of the null hypothesis. This confirms that even for the Feynman dataset's functions, curvature-based adaptation provides a statistically significant advantage. Finally, the computational overhead is consistent to previous results, with average training times of 15.36s for the input-based method versus 17.62s for the curvature-based approach.

\subsection{Benchmark III: Helmholtz Equation} \label{sec4.3}

To evaluate the proposed framework beyond standard regression, we consider the domain of PIML, where KANs have demonstrated significant potential \cite{PINNReview}. Specifically, we solve the forward problem for the 2D Helmholtz equation -- a standard benchmark that presents non-trivial challenges compared to simpler problems like the 1D heat equation, while remaining solvable without adaptive training techniques \cite{rgakans}. 

We consider the domain $\Omega = [-1, 1]^2$ and seek the solution $u(x, y)$ satisfying:

\begin{align}
	\Delta u + k^2(x,y) u &= f(x,y), & \quad (x,y) \in \Omega, \label{eq:helmholtz} \\
	u(x,y) &= 0, & \quad (x,y) \in \partial \Omega, \label{eq:bc}
\end{align}

\noindent where $f(x,y)$ is the forcing term corresponding to the true solution $u_{true}(x,y) = \sin(a_1 \pi x) \sin(a_2 \pi y)$, and $k(x,y) = 1$. We investigate the impact of grid adaptation across four distinct parameter configurations representing increasing frequency: $(a_1, a_2) \in \{(1,1), (1,2), (2,2), (2,4)\}$\footnote{Preliminary experiments indicated that for higher frequencies, e.g., $(4,4)$, training tends to diverge regardless of the grid adaptation strategy employed.}.

\begin{figure}[t!]
	\centerline{\includegraphics[width=\linewidth]{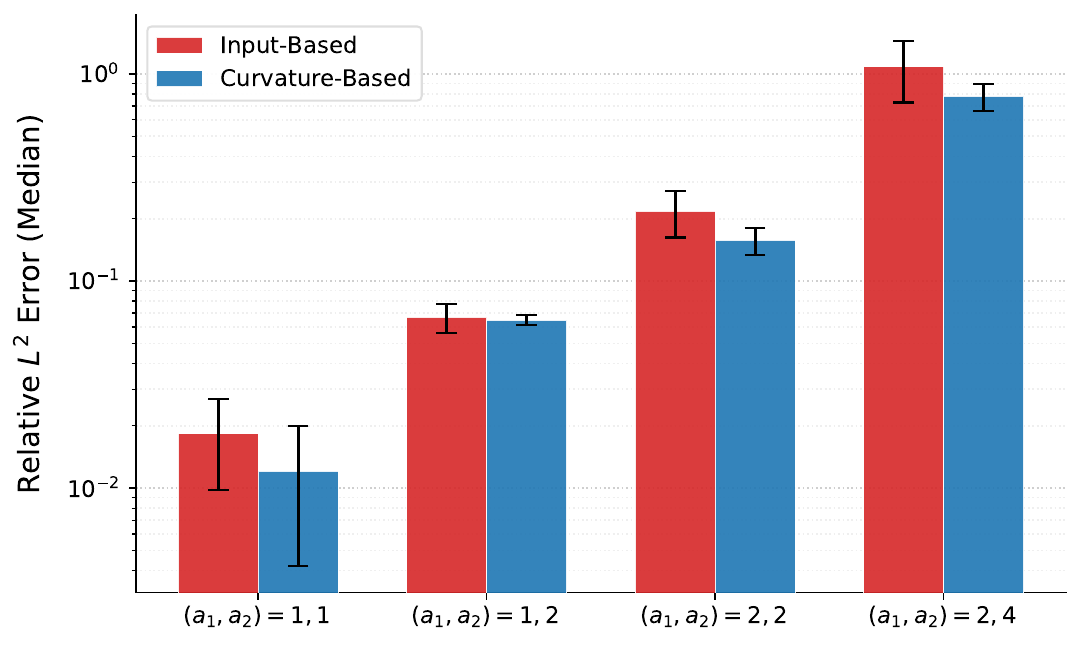}}
	\caption{Comparative evaluation on the Helmholtz PDE benchmark across four increasing frequency configurations. The grouped bar chart showcases the median relative $L^2$ error in logarithmic scale, with error bars indicating the standard deviation over 3 independent runs.}
	\label{fig:pde_results}
\end{figure}

For training, we follow the standard procedure of minimizing a composite physics-informed loss function comprising a PDE residual and a boundary condition term \cite{PIML}. To this end, we utilize a uniform grid of $N_c = 64 \times 64 = 4096$ collocation points to enforce the PDE residual and $N_b = 64$ points along each of the four edges ($256$ total) to enforce the Dirichlet boundary conditions. We employ a deeper KAN architecture with two hidden layers of width 6 and train model instances for 5000 iterations with a learning rate of $\eta = 10^{-3}$. The grid extensions and subsequent adaptations are performed at iterations 1000, 2000, and 3000. The quantitative results on this benchmark are visualized in the grouped bar chart of Figure \ref{fig:pde_results}, where the bars represent median $L^2$ error relative to the reference solution $u_{true}$ and the error bars correspond to the standard deviation.

Consistent with the previous benchmarks, the curvature-based strategy outperforms the input-based baseline across all investigated parameter configurations. While the log scale of Figure \ref{fig:pde_results} visually compresses the differences, the quantitative gains are substantial: the proposed method yields relative improvements of 34.20\% for $(1,1)$, 3.15\% for $(1,2)$, 27.65\% for $(2,2)$, and 28.10\% for the high-frequency case $(2,4)$. This corresponds to an average reduction in error of 23.27\%.

Besides the improvement in median accuracy, we again observe that the curvature-based results exhibit lower variance across independent seeds compared to the baseline. This reinforces the hypothesis that curvature-based adaptation leads to more robust training outcomes. Due to the small sample size ($N=4$ configurations), a Wilcoxon signed-rank test is statistically irrelevant; although the curvature method outperforms the baseline in every single instance (yielding the minimum possible $p$-value of $0.0625$), this falls just above the substantial threshold of $0.05$. Nonetheless, the consistent superiority across all tests, combined with the clear physical intuition -- that knots should cluster in regions of high wavefront curvature rather than being distributed uniformly -- strongly supports the efficiency of the method.

Finally, we note that as the complexity of the training task increases, the relative computational cost of the proposed method diminishes. For this benchmark, which involves deeper networks and 5000 training iterations, the average wall-clock time is 61.21s for the input-based method and 64.30s for the curvature-based approach. This represents an overhead of only $\approx 5\%$, significantly lower than in previous benchmarks. This trend suggests that for large-scale scientific machine learning tasks involving prolonged training, the additional cost of computing training dynamics becomes negligible relative to the total training time, rendering the two methods practically equivalent in terms of computational expense.

\section{Conclusion \& Outlook} \label{sec5}

In this work, we introduced a generalized framework for grid adaptation in Kolmogorov--Arnold Networks, re-framing the problem of knot allocation as a density estimation task governed by Importance Density Functions. We demonstrated that the standard input-based adaptation strategy employed in existing KAN implementations is a special case of this framework corresponding to a uniform IDF. We proposed a curvature-based adaptation strategy that aligns grid resolution with the geometric complexity of the target function. Through experiments on regression and physics-informed PDE solving, we showed that this approach consistently outperforms the standard input-based baseline, achieving significant error reductions and improved training stability with negligible computational overhead.

While our results are promising, several limitations of this study warrant discussion. First, this work primarily established the theoretical framework and evaluated a single alternative IDF (curvature) against the standard input-based method. While curvature is geometrically motivated, it represents only one possible heuristic for importance. Moreover, our evaluation was concentrated on the domain of scientific machine learning. While KANs have shown particular promise in this field, they are a fundamental neural architecture applicable to broader domains. We did not evaluate our method on tasks with high-dimensional data such as audio or image processing, partly because KANs have yet to establish state-of-the-art performance in these areas comparable to architectures like Transformers. Finally, the experiments conducted here utilized relatively compact architectures (1-2 hidden layers) appropriate for the regression and PDE tasks at hand. As evidenced by the low relative errors achieved (orders $10^{-2}$ to $10^{-4}$ for regression), these were sufficient. However, it remains to be seen how the proposed method scales to significantly deeper, parameter-heavy architectures required for more complex tasks, although the observed trend of diminishing relative computational cost suggests favorable scaling.


The proposed framework opens several avenues for future research. A direct extension involves the design and evaluation of novel IDFs; for instance, importance could be derived from per-sample loss values, or via higher-order derivative metrics as suggested by approximation theory. Another promising direction lies in the timing of grid updates. Currently, they occur at fixed, predetermined intervals. A more automated approach, similar to what is proposed in the AdaptKAN framework \cite{adaptKAN}, could involve defining ``trigger'' metrics monitored during training -- such as plateaus in the loss landscape or spikes in validation error -- to automatically initiate grid adaptation or extension steps. This would move KANs towards fully autonomous self-adaptation, further reducing the need for manual hyperparameter tuning in scientific computing workflows.

\section*{AI Usage Statement}

Large Language Models (LLMs) were used throughout this work for grammar and syntax refinement only; all ideas, technical content and conclusions remain the authors' work.

\section*{Reproducibility Statement}

The full code (including selected seeds for each experiment) and the processed data are publicly available at {\small \texttt{https://github.com/srigas/kan\_grid}}.

\section*{Acknowledgment}

This work was supported by the Greek CRE project ``ThermaGraph'' (ΥΠ3ΤΑ-0559393).

\bibliographystyle{IEEEtran}
\bibliography{IEEEabrv,references}

\end{document}